\documentclass[conference]{IEEEtran}

\usepackage{cite}
\usepackage{amsmath,amssymb,amsfonts}
\usepackage{graphicx}
\usepackage{textcomp}
\usepackage[table]{xcolor}
\usepackage{booktabs}
\usepackage{multirow}
\usepackage{graphicx}
\usepackage{subcaption}

\def\BibTeX{{\rm B\kern-.05em{\sc i\kern-.025em b}\kern-.08em
    T\kern-.1667em\lower.7ex\hbox{E}\kern-.125emX}}

\title{Rethinking Speech-LLM Integration for ASR: Effective Joint Speech-Text Training by Interleaving}

\author{
\IEEEauthorblockN{Ruchao Fan, Yiming Wang, Rui Zhao, Liliang Ren, Keqi Deng, Xiaoyang Chen\\ 
Ali Zare, Bo Ren, Yuxuan Hu, Junkun Chen, Yan Huang, Yelong Shen, Jinyu Li}
\IEEEauthorblockA{\textit{Microsoft, USA}\\
}
}

\begin{document}
\maketitle

\begin{abstract}
    Speech-LLM integration has shown promising results by leveraging extensive textual pretraining, yet its specific benefits for automatic speech recognition (ASR) remain unclear. We observe that as supervised ASR training data increases, the contribution of LLM priors becomes less evident, and simple speech-text joint training under-utilizes textual knowledge. We therefore propose Joint Speech-Text Interleaved Pretraining (JSTIP), an ASR-oriented pretraining strategy that constructs word-level and segment-level interleaved speech-text sequences within aligned pairs for speech-LLM architectures that accept continuous inputs. Experiments on 38k hours of ASR data show consistent entity accuracy improvement compared to ASR-only and joint speech-text training baselines. JSTIP achieves on-par entity recognition performance using domain transcription text compared to synthetic speech-text pairs, simplifying domain adaptation. Benefiting from textual pretraining and domain text data, JSTIP is competitive with open-source ASR and Speech-LLM systems in medical entity recognition. The zero-shot speech question answering behaviors further suggest that interleaving reduces the speech-text modality gap and preserves the LLM generative prior, which is likely the reason for the entity improvements on the ASR task.
    
\end{abstract}


\section{Introduction}
\label{sec:intro}

\begin{figure*}[th!]
    \centering
    \includegraphics[width=0.98\textwidth,height=0.43\textwidth]{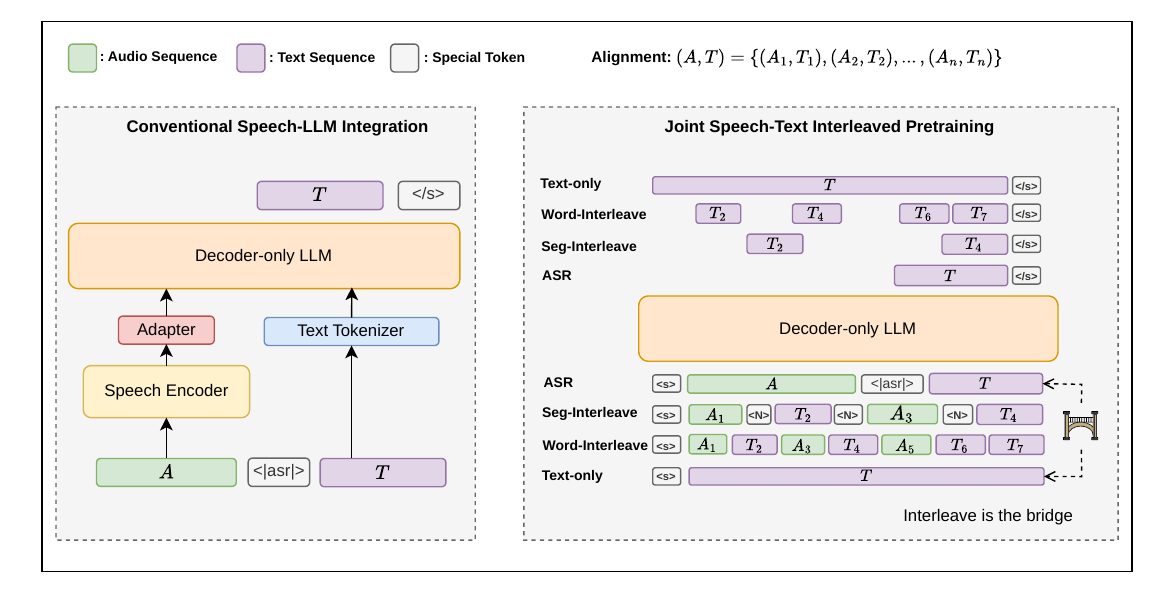}
    \caption{Frameworks for conventional Speech-LLM integration (left) and joint speech-text interleaved pretraining (right). }
    \label{fig:jstip}
\end{figure*}

Large language models (LLMs) pretrained on massive text corpora exhibit strong linguistic modeling capability and broad world knowledge~\cite{achiam2023gpt,touvron2023llama,liu2024deepseek,team2025gemma,yang2025qwen3}. Motivated by the strengths, recent studies integrate LLMs into speech systems under a decoder-only architecture~\cite{wu2023decoder,defossez2024moshi,tang2024salmonn,abouelenin2025phi,ding2025kimi,wu2025step,xu2025qwen3, hu2026slms2stmultimodallanguagemodel, team2025fun}. A common approach connects a pretrained speech encoder to an LLM via an adapter and performs multi-stage training, where automatic speech recognition (ASR) data is used to align speech and text representations. The expectation for ASR falls into two aspects. First, LLM pretraining is assumed to improve ASR performance by compensating for the limited size and coverage of supervised ASR data, particularly for rare words, domain-specific terminologies, and long-tail entities. Indeed, LLM-based ASR systems have shown promising results~\cite{xu2025fireredasr,bai2024seed,an2025fun,song2025index,ma2025speech,mu2025efficient,shi2026qwen3,deng2025transducer,saon2025granite,shi2025train,ren2026rlbr}. Second, properly aligned speech–LLM models with ASR are expected to inherit zero-shot instruction-following capability from the backbone LLM~\cite{deng2024wav2prompt,fan2025alignformer}. However, under conventional training paradigms, ASR performance does not consistently benefit from pretrained textual knowledge. As ASR training data increases, the accuracy is largely governed by speech–text supervision, and entity improvements tend to correlate with ASR data coverage rather than LLM priors. Even when additional text data is included, ASR often struggles to effectively exploit the LLM backbone or text data. 


In this work, we revisit speech–LLM integration from a pretraining perspective. 
When ASR supervision dominates, the decoder specializes in speech modeling and weakens the generative prior learned during text pretraining. We show that this issue persists even in standard speech–text joint training if text-only modeling behavior is not explicitly preserved. Therefore, we propose \textbf{Joint Speech–Text Interleaved Pre-Training (JSTIP)} for decoder-only architectures that emphasizes interleaved sequence construction within aligned speech–text pairs. Word-level and segment-level alignments are obtained initially, and speech and text segments are interleaved to form training sequences that alternate between modalities. This design maintains consistent modeling behavior across speech–text and text-only sequences, thereby reducing the modality gap and preserving the LLM's generative prior. 


Experiments on 38k hours of in-house ASR data show that JSTIP achieves up to 17.2\% relative entity improvement over ASR-only training. The gains are consistent with more effective transfer of LLM knowledge and domain text data into ASR. In a domain-specific setting, JSTIP achieves similar entity performance using transcription-only domain text compared to synthetic domain ASR data. The best JSTIP system is competitive with open-source ASR and Speech-LLM systems in medical entity performance. Moreover, JSTIP enables zero-shot speech question answering (SQA) capability without introducing additional task-specific supervision. We use SQA as evidence that the model preserves text-side generative behavior, rather than as a separate task focus.



\section{Related Work}
\label{sec:related_work}

\subsection{Speech-LLM Integration}
\label{ssec:related_work:speech-llm-integreation}
Speech-LLM integration has been widely explored recently. Representative systems include the Qwen-audio series~\cite{xu2025qwen2,xu2025qwen3}, Phi-4-Multimodal~\cite{abouelenin2025phi}, Moshi~\cite{defossez2024moshi}, Step-Audio~\cite{wu2025step}, Kimi-Audio~\cite{ding2025kimi}, and Voxtral~\cite{liu2025voxtral}. These systems commonly connect a speech encoder to a decoder-only LLM and train on speech and text data. Several technical reports indicate that mixed or interleaved speech-text pretraining is useful for general multimodal ability, but the precise data construction and its ASR-specific effect are often not isolated. As a result, it remains unclear when LLM text pretraining improves ASR once substantial supervised ASR data is already available.

Speech-text interleaving has also been studied in other Speech-LLM settings~\cite{seide2024speech,zeng2024scaling}. Spirit-LM~\cite{nguyen2024spiritlm} interleaves discrete speech and text tokens. Voxtral~\cite{liu2025voxtral} and Xie et al.~\cite{xie2025enhancing} use segment-level constructions. Our study does not claim to introduce interleaving itself; it contributes scalable word-level interleaving for continuous representations, mixed word/segment construction, and controlled ASR evaluation at 38k-hour scale. Word-level interleaving requires variable-length aligned acoustic spans to be encoded, reinserted into the LLM stream, and packed efficiently. Our analysis further isolates entity recognition and transfer from domain text rather than general instruction-following or speech generation alone.

Therefore, our work differs from existing Speech-LLM integration studies in three central aspects. First, we study interleaving as a mechanism for preserving and transferring the pretrained LLM prior to ASR, rather than as a general multimodal data recipe. Second, we focus on the large-supervised-data regime where conventional ASR adaptation already learns strong speech-text mapping and where the additional value of LLM pretraining is hardest to expose. Third, we isolate how construction granularity affects the modality gap and entity recognition, including a scalable word-level interleaving implementation for continuous speech representations that is not covered by prior segment-level or discrete-token formulations.

\subsection{Joint Speech and Text Training}
\label{ssec:related_work:joint-speech-text}

Prior studies inject text-only data into speech training to exploit linguistic regularities and long-tail knowledge absent from paired data. Early efforts integrated text into self-supervised pretraining, either by synthesizing speech from text~\cite{chen2021injecting} or learning shared speech–text representations~\cite{zhang2022speechlm,Ao2021SpeechT5UE,bapna2021slam}. However, multi-stage training often led to forgetting during downstream ASR fine-tuning~\cite{sainath2023joist,bai2022joint,10447838}. More direct text injection was later explored within AED and transducer frameworks~\cite{sainath2023joist,DBLP:conf/interspeech/WangSW21,sainath2020attention,Chen2022MAESTROMS,peyser23_interspeech}, where text-only inputs were incorporated during supervised ASR optimization. Because these models decode conditioned on text histories, various alignment strategies, such as duration-based up-sampling~\cite{Chen2022MAESTROMS} or repetition heuristics~\cite{sainath2023joist}, were required to mitigate the modality mismatch between text and speech input.

While effective, these methods were developed primarily in the pre-LLM era to enhance in-domain recognition. In modern speech–LLM systems, the challenge is no longer merely strengthening a decoder language model, but preserving and transferring the extensive textual prior already encoded in a pretrained LLM during multimodal adaptation. As we show in this work, naive combination of speech and text objectives improves text-side prediction but transfers only weakly to speech-conditioned ASR. JSTIP revisits text injection under the decoder-only LLM paradigm, with a particular emphasis on maintaining consistent modeling behavior across speech–text and text-only sequences. This positioning separates JSTIP from prior joint speech-text training. Instead of using text as an auxiliary data source for a separately trained ASR decoder, JSTIP changes the sequence construction inside aligned speech-text pairs so that the same decoder repeatedly predicts text after both speech and text contexts. This design directly addresses the modality gap introduced during Speech-LLM adaptation and provides a route for domain transcription text to improve speech-conditioned entity recognition without requiring synthetic speech for every text example.

\section{Method}
\label{sec:method}


\subsection{Conventional Speech-LLM Integration}
\label{ssec:method:speech-llm-training}

Decoder-only LLMs unify language tasks under the next-token prediction (NTP) paradigm. For speech understanding, a typical integration pipeline consists of three components: a pretrained speech encoder, a modality adapter, and a pretrained decoder-only LLM, as shown in the left in Figure.~\ref{fig:jstip}. Given an input audio sequence $A$, the speech encoder produces a sequence of acoustic representations and an adapter projects the encoder output into the LLM embedding space.
The projected acoustic representations are concatenated with special tokens (e.g., \texttt{<|asr|>}) and fed into the decoder-only LLM to generate the target transcription sequence $T = (t_1, \dots, t_N)$.

\noindent\textbf{ASR-only Training}: Under traditional ASR training, the model is optimized using cross-entropy loss over the target:

\begin{equation}
\mathcal{L}_{\text{ASR}}
=
- \sum_{i=1}^{N}
\log P(t_i \mid A, t_{<i}),
\label{eq:asr_loss}
\end{equation}
where $t_{<i}$ denotes previously generated tokens. This formulation treats ASR as a speech-conditioned NTP problem. As the amount of supervised ASR data increases, optimization is dominated by speech--text supervision, which may reduce reliance on textual priors learned during LLM pretraining.

\noindent\textbf{Speech-Text Joint Training}: with decoder-only architecture, it is easy to include text-only data in the training by creating zero speech input for speech encoder when text batch comes. Consequently, the text data can contribute to the gradient updates for LLM, biasing towards to the target ASR domain. However, we find that a simple mixing of the data types cannot transfer the text knowledge into ASR performance.

\subsection{Joint Speech-Text Interleaved Pretraining}

To mitigate the over-specialization toward speech-conditioned decoding, we study Joint Speech--Text Interleaved Pretraining (JSTIP). Unlike dataset-level scheduling between ASR and text-only data, JSTIP constructs interleaved sequences \emph{within each aligned speech-text pair}. Given a speech-text pair $(A, T)$, we first obtain the alignment information $(A, T) = \{(A_1, T_1), (A_2, T_2), \dots, (A_n, T_n)\}$, where each $A_i$ corresponds to the acoustic segment aligned with text segment $T_i$. 
With the alignment, we construct interleaved speech-text sequences by alternatively selecting speech and text segments.
Two complementary variants are used $(A_1, T_2, A_3, T_4, \dots, T_n)$ and $(T_1, A_2, T_3, A_4, \dots, T_n)$.
In both variants, the final segment of the sequence is text, and the text segments serve as the training target. We use deterministic alternation to isolate the impact of interleaving granularity and boundary construction; adaptive choices such as random or word-type-based modality selection are left for future work. Based on the granularity of the alignment, word-level and segment-level interleaved sequences are studied in this work. 

\noindent\textbf{Loss Masking}: For all training formats, cross-entropy loss is applied only to text tokens. In ASR-only training, the loss is computed on transcript tokens; in text-only training, it is computed on text tokens; and in interleaved training, it is computed on all text tokens that appear in the constructed interleaved sequence. Speech positions are always ignored by the loss mask. Special tokens are used only as task or modality indicators and are excluded from the loss. When multiple examples are packed into the same 8k context, we use packed SFT with cumulative sequence lengths (\texttt{cu\_seqlens}) in FlashAttention so that attention is reset at example boundaries and no cross-example leakage occurs.

Finally, ASR-only, segment-interleave, word-interleave, and text-only data are trained under a unified next-token prediction objective. Unlike conventional ASR training in Eq.~\ref{eq:asr_loss}, JSTIP exposes the model to distributions of the form:

\begin{equation}
P(T_j \mid A_i, T_{i+1}, \dots),
\end{equation}
thereby preserving textual modeling behavior while enabling knowledge transfer from text data into speech tasks. Interleaving thus acts as a structural bridge between modalities, ensuring that the decoder-only LLM retains its generative prior during multimodal adaptation.

\noindent\textbf{Word-Level Interleave}: alignment is performed at fine granularity. Each $T_i$ corresponds to a word (or token span), and $A_i$ represents its aligned acoustic region. The constructed sequence alternates more frequently. We skip any special tokens between the speech and text segments for the word-level interleaving, enforcing tighter cross-modal coupling and reducing the modality gap at a finer scale. Word-level interleaving is challenging when continuous speech tokens are used as input in pretraining, because processing many short acoustic spans independently creates large zero-padding overhead in the speech encoder and can make large-batch training run out of GPU memory. To mitigate this issue, we concatenate all speech segments into a longer sequence at the input level and place them back into their original interleaved positions after the adapter forward computation, making word-level interleaving feasible at scale while reducing GPU memory usage. We find that this strategy improves speech-text joint training. 

\noindent\textbf{Segment-Level Interleave}: each pair $(A_i, T_i)$ corresponds to a phrase-level or sentence-level segment. The constructed sequence can be written as:

\begin{equation}
\langle s \rangle, A_1, \langle N \rangle, T_2, \langle N \rangle, A_3, \langle N \rangle, T_4, \dots, \langle N \rangle, T_n, \langle /s \rangle,
\end{equation}
where $\langle s \rangle$ and $\langle /s \rangle$ denote start and end tokens. $\langle N \rangle$ is a leading token to differentiate with the ASR task.

\section{Experimental Setup}
\label{sec:experimental_setup}

All experiments are conducted under a unified LLM pretraining framework with 8k context length.

\subsection{Training Data}
\label{ssec:training-data}
\textbf{ASR Data}: We use 38k hours of anonymized English in-house ASR data as the primary source for ASR training. The 38k hours correspond to approximately 2.3B training tokens with a token rate of 12.5HZ for speech-LLM. Evaluation speakers are disjoint from training speakers. To further validate the effectiveness of JSTIP in domain adaptation setup, we synthesize 9k hours of medical-domain ASR data (TTS-pairs). The transcriptions are created by prompting GPT for entity-rich utterances across medical topics, and the audio is generated by an in-house TTS model.

\noindent\textbf{Interleaved Data}: To construct interleaved training sequences, we first obtain word-level alignment on the 38k-hour ASR data using an HMM-based hybrid ASR system. Utterances with failed alignments are removed, while no additional filtering is applied based on alignment confidence. Based on the alignments, two types of interleaving units are derived and studied. In word-level interleave, each aligned word forms a minimal unit. In the segment-level interleave, consecutive words are merged into segments according to either acoustic (silence) or text (punctuation) signal. Each JSTIP run uses 2.3B interleaved tokens, formed as deterministic views of the same aligned ASR pairs; these views add no source utterances, labels, or external knowledge. Alignment quality can affect word-level interleaving because boundary errors directly change the speech spans paired with words; segment-level interleaving is expected to be less sensitive to such local errors. Robustness to alternative aligners, such as CTC alignments or timestamp-based forced alignment, is left for future work.

\noindent\textbf{Text Data}: To preserve textual modeling capacity, we incorporate text-only corpora, including 2.3B text tokens sourced from PubMed~\footnote{https://pubmed.ncbi.nlm.nih.gov/} abstracts and 0.1B text tokens from the transcripts of TTS-pairs. The text data are cleaned and prepared into the LLM pretraining format truncated at 8k context length.

\subsection{Model and Training}
The model follows a decoder-only speech–LLM architecture. The speech encoder is a 400M-parameter Conformer with a temporal downsampling factor of 8. Input features are 80-dimensional log Mel filterbanks extracted with a 10ms frame shift, resulting in an effective 80ms token rate for the decoder.

The decoder is initialized from an internal 7B-parameter LLM, which was trained with 5T text tokens. A lightweight ($\sim$20M) adapter projects encoder outputs into the LLM latent space. Training is performed in two stages. In the first stage, only the adapter parameters are updated using ASR data to stabilize cross-modal alignment. The learning rate is set to $1\times10^{-4}$ with linear decay using AdamW, and training covers 10\% of the total ASR tokens. In the second stage, all parameters—including encoder, adapter, and LLM—are jointly optimized using a learning rate of $4\times10^{-5}$ with linear decay and AdamW optimization. Both the ASR and interleaved sequences are packed into 8k training context length. We use packed SFT with sequence-boundary information in FlashAttention and loss masks to avoid cross example leakage. Each included corpus is traversed once in Stage 2. Thus, paired rows hold source-data coverage fixed, while JSTIP uses additional compute for the deterministic interleaved view. With a 256K-token global batch, standard/JSTIP training uses 2.3B/4.6B tokens (8.8K/17.5K updates) for ASR, and increased token and update budgets correspondingly. The composite system uses 7.46B tokens (28.5K updates).
\begin{table*}[t]
\renewcommand{\arraystretch}{1.03}
\setlength{\tabcolsep}{3.2pt}
\centering
\caption{Overall results for JSTIP. T2T is text-to-text prediction and S2T indicates speech-in text-out prediction. TTS-pairs are synthetic medical-domain ASR data, while TTS-transcription is their text. +Interleave is the best mixed word/segment JSTIP variant. Paired rows fix source corpora; +Interleave uses one additional deterministic view and therefore more compute.}
\label{tab:ASR_overall}
\resizebox{\textwidth}{!}{
\begin{tabular}{l ccc cc cc cc ccc}
\toprule
& \multicolumn{3}{c}{Data Type}
& \multicolumn{2}{c}{TER $\downarrow$}
& \multicolumn{2}{c}{EER $\downarrow$}
& \multicolumn{2}{c}{ACC $\uparrow$ (5-shot) }
& \multicolumn{2}{c}{ACC $\uparrow$ (0-shot) } \\
\cmidrule(lr){2-4}\cmidrule(lr){5-6}\cmidrule(lr){7-8}\cmidrule(lr){9-10}\cmidrule(lr){11-12}
Model 
& ASR & Interleave & Text & Conversation & Dictation & Medical-AVG & Banking & MMLU-T2T & MMLU-S2T &  SQA-T2T & SQA-S2T \\ 
\midrule
LLM-7b & $\times$ & $\times$ & $\checkmark$ & - & - & - & - & 78 & - & 60.17 & - \\ 
\midrule
ASR-only & $\checkmark$ & $\times$ & $\times$                                          & 23.63 & 11.06 & 7.97 & 11.57 & 43.01 & 35.68 & 9.41 & 0.05 \\
\rowcolor[gray]{.8} \hspace{1pt} +Interleave & $\checkmark$ & $\checkmark$ & $\times$     & 22.65 & 10.81 & 7.32 & 11.29 & 51.26 & 51.77 & 45.37 & 41.92 \\

ASR+PubMed & $\checkmark$ & $\times$ & $\checkmark$                                    & 23.32 & 10.81 & 7.49 & 11.29 & 64.1 & 43.77 & 43.97 & 10.1 \\

\rowcolor{green!30} \hspace{1pt} +Interleave & $\checkmark$ & $\checkmark$ & $\checkmark$ & 22.35 & 10.74 & 6.87 & 9.98 & 64.16 & 58.98 & 44.95 & 41.03 \\
\midrule
ASR+TTS-pairs & $\checkmark$ & $\times$ & $\times$                                     & 22.71 & 10.73 & 6.86 & 10.8 & 48.17 & 36.96 & 36.47 & 7.6 \\
\rowcolor[gray]{.8} \hspace{1pt} +Interleave & $\checkmark$ & $\checkmark$ & $\times$     & 22.22 & 10.54 & 6.72 & 10.53 & 54.98 & 53.89 & 44.59 & 41.12 \\
ASR+TTS-transcription & $\checkmark$ & $\times$ & $\checkmark$                         & 23.69 & 11.27 & 7.85 & 11.02 & 54.44 & 37.95 & 42.63 & 6.27 \\
\rowcolor{green!30} \hspace{1pt} +Interleave & $\checkmark$ & $\checkmark$ & $\checkmark$ & 22.3 & 10.49 & 6.81 & 10.47 & 57.02 & 53.26 & 43.81 & 40.39\\
\midrule
\multicolumn{12}{c}{Composite: ASR + Interleave + TTS-Interleave + PubMed + TTS-transcription} \\
\midrule
\rowcolor{green!30} JSTIP-Best-EER & $\checkmark$ & $\checkmark$ & $\checkmark$                            & 22.42 & 10.48 & \textbf{6.60} & 10.75 & 64.09 & 58.7 & 44.92 & 42.07 \\ 

\bottomrule
\bottomrule
\end{tabular}}
\end{table*}

\begin{table*}[t!]
\renewcommand{\arraystretch}{1.03}
\setlength{\tabcolsep}{3.0pt}
\centering
\caption{Comparisons to open-source models on domain sets. The numbers are entity error rates (EER, \%). Medical-AVG is averaged over eight held-out entity-rich medical test sets. JSTIP (ours) uses 38k ASR data only while other models might use millions of hours of ASR training data.}
\label{tab:compare_to_oss}
\resizebox{\textwidth}{!}{
\begin{tabular}{l cccccccc}
\toprule
\multirow{2}{*}{EER/Model} & Whisper & \multicolumn{3}{c}{Qwen} & \multicolumn{2}{c}{Voxtral} & Gemma & JSTIP \\
\cmidrule(lr){3-5}\cmidrule(lr){6-7}
~ & Large-V3 & ASR-1.7B & Omni-2.5-7B & Omni-3-30BA3B & Mini-3B & Small-24B & 3n-E4B & (Ours) \\
\midrule
\rowcolor{green!30} \textbf{Medical-AVG} & 6.94 & 6.67 & 12.22 & \textbf{5.84} & 7.40 & 6.04 & 10.62 & 6.60 \\
Banking & \textbf{8.88} & 9.81 & 19.13 & 9.87 & 10.25 & 9.38 & 17.21 & 10.75 \\
AVG-All & 7.16 & 7.02 & 12.99 & \textbf{6.29} & 7.71 & 6.41 & 11.35 & 7.06 \\

\bottomrule
\bottomrule
\end{tabular}}
\end{table*}


\subsection{Evaluation}
We evaluate the model via general ASR, domain-specific entity accuracy, and speech question answering (SQA).

\noindent\textbf{ASR and Entity Evaluation}: For general ASR evaluation, we use two held-out in-house test sets, \emph{conversation} and \emph{dictation}, measured by Token Error Rate (TER), which counts all tokens including capitalization and punctuation. The conversation set reflects spontaneous short-form speech, while the dictation set reflects longer-form dictated speech with richer punctuation and formatting. For domain evaluation, we use a selected suite of 9 held-out in-house test sets covering eight medical domains and one banking domain, which are designed to be entity-rich, with references and entity spans human annotated from the transcripts. The medical aggregate and banking set contain 49.3/5.6 hours of speech, and 31k/1.8k entities, respectively. Performance is measured by Entity Error Rate (EER), computed as $1 - \mathrm{recall}$ over annotated entity instances. EER is computed before text normalization, so formatting, capitalization, punctuation, and spelling are valued; spelling variants are not accepted. We use EER rather than domain TER as the primary metric because the effect of LLM knowledge transfer is concentrated on entities, while aggregate TER is dominated by common non-entity tokens and showed only modest differences in our development analysis. 

\noindent\textbf{Open-Source Baselines}: For the open-source comparisons in Table~\ref{tab:compare_to_oss}, all models are evaluated on the same audio and scored by the same EER pipeline. We use each model's official inference implementation with the prompt and decoding settings recommended by the corresponding model webpage. Because these systems differ in model size, training data, prompt format, and decoding recipe, the comparison is intended as an external reference rather than a fully controlled ranking of Speech-LLM architectures.

\noindent\textbf{Next Token Prediction}: To quantify the modality gap, we construct a speech version of MMLU~\cite{hendrycks2020measuring} evaluated in a 5-shot setting. Performance is measured under both text-to-text (T2T) and speech-to-text (S2T) conditions. Answers are predicted via next-token selection over options \{A, B, C, D\}, with random guess accuracy at 25\%. The performance difference between S2T and T2T reflects the extent to which textual reasoning behavior is preserved after speech adaptation.

\noindent\textbf{Zero-shot Speech Question Answering (SQA)}: Finally, we assess the zero-shot SQA capabilities of our model on three datasets: LLaMA-QA, TriviaQA, and WebQA, which constitute a subset of UltraEval-Audio~\cite{shi2026ultraeval}. We report the average accuracy across the three sets. Unlike MMLU that evaluates next-token prediction consistency, SQA requires open-ended answer generation and therefore serves as an additional diagnostic for whether text-side generative behavior remains accessible under speech input. SQA can also be affected by prompting, decoding, and answer normalization, so we interpret it as supporting evidence for knowledge preservation rather than as a standalone explanation for entity gains.

\section{Results}
\label{sec:results}

\subsection{Main Results on ASR}
Table~\ref{tab:ASR_overall} summarizes the performance of JSTIP under different configurations. We report TER on conversation and dictation sets, EER on medical and banking domain sets, and accuracy on MMLU and SQA benchmarks. TER changes are modest, which is expected under 38k hours of supervised ASR training; the main gains appear on entity recognition, where external textual knowledge is most useful. MMLU and SQA are used as diagnostics for whether the speech-conditioned model can still access the LLM's text-side generative prior.

Without additional domain text, interleaving improves both entity recognition and zero-shot SQA capability. The averaged EER on 8 medical domain sets decreases from 7.97\% to 7.32\%, and SQA-S2T accuracy increases from 0.05\% to 41.92\%, suggesting that interleaving helps preserve next-token prediction (NTP) behavior inherited from LLM pretraining. This helps explain why text pretraining can appear less effective under large ASR supervision: ASR-only adaptation leaves MMLU-S2T at 35.68\% even though the LLM was pretrained on 5T text tokens. When domain text is added without interleaving (ASR+PubMed), MMLU-T2T improves substantially (43.01 $\rightarrow$ 64.1) because text sequences update the LLM, but MMLU-S2T and EER improve much less, showing that the modality gap prevents text-side gains from fully transferring to speech. Incorporating interleaving in the same setting raises MMLU-S2T to 58.98\%, reduces medical EER to 6.87\%, and increases SQA-S2T to 41.03\%, suggesting more effective utilization of textual priors for ASR.

\begin{table*}[t!]
\centering
\caption{Ablation study on the interleave sequence (IL). Word-level alignment is merged into segments based on acoustic signal (silence) and/or semantic signal (punctuation). TTS-trans. indicates the transcription of the synthetic domain ASR data.}
\label{tab:ablation_segment}
\resizebox{0.78\textwidth}{!}{
\begin{tabular}{l ccccc}
\toprule
Interleave Type & Strategy & EER-Medical & MMLU-Text & MMLU-Speech & $\Delta$ (S2T$-$T2T) \\ 
\midrule
ASR only & - & 7.97 & 43.01 & 35.68 & -7.33 \\
\midrule
  \hspace{1pt} + Word-IL & - & 7.64 & 46.83 & 39.94 & -6.89 \\
\midrule
\multirow{2}{*}{\hspace{1pt} + Segment-IL} & silence & 7.79 & 54.37 & 49.92 & -4.45 \\
~ & sil. + punc. & 7.69 & 53.59 & 52.82 & \textbf{-0.77} \\
\midrule
\multirow{2}{*}{\hspace{1pt} + Mixed-IL} & silence & 7.55 & 56.27 & 52.16 & -4.11   \\
~ & sil. + punc. & \textbf{7.32} & 51.26 & 51.77 & \textbf{+0.61} \\
\midrule\midrule
ASR+TTS-pair & - & 6.86 & 48.17 & 36.96 & -11.21 \\
ASR+TTS-trans. & - & 7.85 & 54.44 & 37.95 & -16.49 \\
\midrule

\multirow{2}{*}{\hspace{1pt} + Mixed-IL} & silence & 7.27 & 58.71 & 55.27 & -3.44   \\
~ & sil. + punc. & \textbf{6.81} & 57.02 & 53.26 & -3.76  \\
\bottomrule
\bottomrule

\end{tabular}}
\end{table*}

We further examine domain transcription data as a cost-efficient alternative to synthetic TTS pairs. Without interleaving, adding TTS-transcription yields negligible EER improvement (7.97\% $\rightarrow$ 7.85\%). With interleaving, however, Medical-AVG EER drops to 6.81\%, achieving performance comparable to synthetic TTS-pair training. This supports the role of interleaving in transferring textual knowledge to ASR. The best configuration (JSTIP-Best-EER), combining 38k hours ASR data, PubMed text, TTS transcription, and TTS-interleaved sequences, achieves the lowest Medical-AVG EER.

Table~\ref{tab:compare_to_oss} further compares JSTIP-Best-EER with open-source ASR and speech-LLM baselines on aggregate medical and banking metrics, including Whisper-large-v3, the Qwen family (Qwen3-ASR-1.7B, Qwen2.5-Omni-7B, and Qwen3-Omni-30BA3B), Voxtral-Mini-3B, Voxtral-Small-24B, and Gemma-3n-E4B. Since the open-source systems differ in model size, training data, prompting, and decoding setup, the table should be interpreted as an external reference comparison. On Medical-AVG, JSTIP-Best-EER trained with 38k hours ASR data outperforms Whisper-large-v3 (6.60\% vs. 6.94\%), Qwen3-ASR-1.7B (6.67\%), Voxtral-Mini-3B (7.40\%), Gemma-3n-E4B (10.62\%), and Qwen2.5-Omni-7B (12.22\%), which are trained with million-hours data. The benefit originates from LLM pretraining and medical domain text data; correspondingly. Since no banking-domain text is used, the banking gains cannot be attributed to PubMed transfer and are instead consistent with improved access to the pretrained LLM's knowledge. The medical average spans eight test sets, reducing sensitivity to any single-domain outlier. JSTIP remains worse than Qwen-30BA3B and Voxtral-24B on the medical aggregate, whose LLM backbones are larger.

\subsection{Ablation Study on Interleaved Sequence}
\label{ssec:results-ablation}

\begin{figure}[t]
    \centering
    \begin{subfigure}{0.48\columnwidth}
        \centering
        \includegraphics[width=\linewidth]{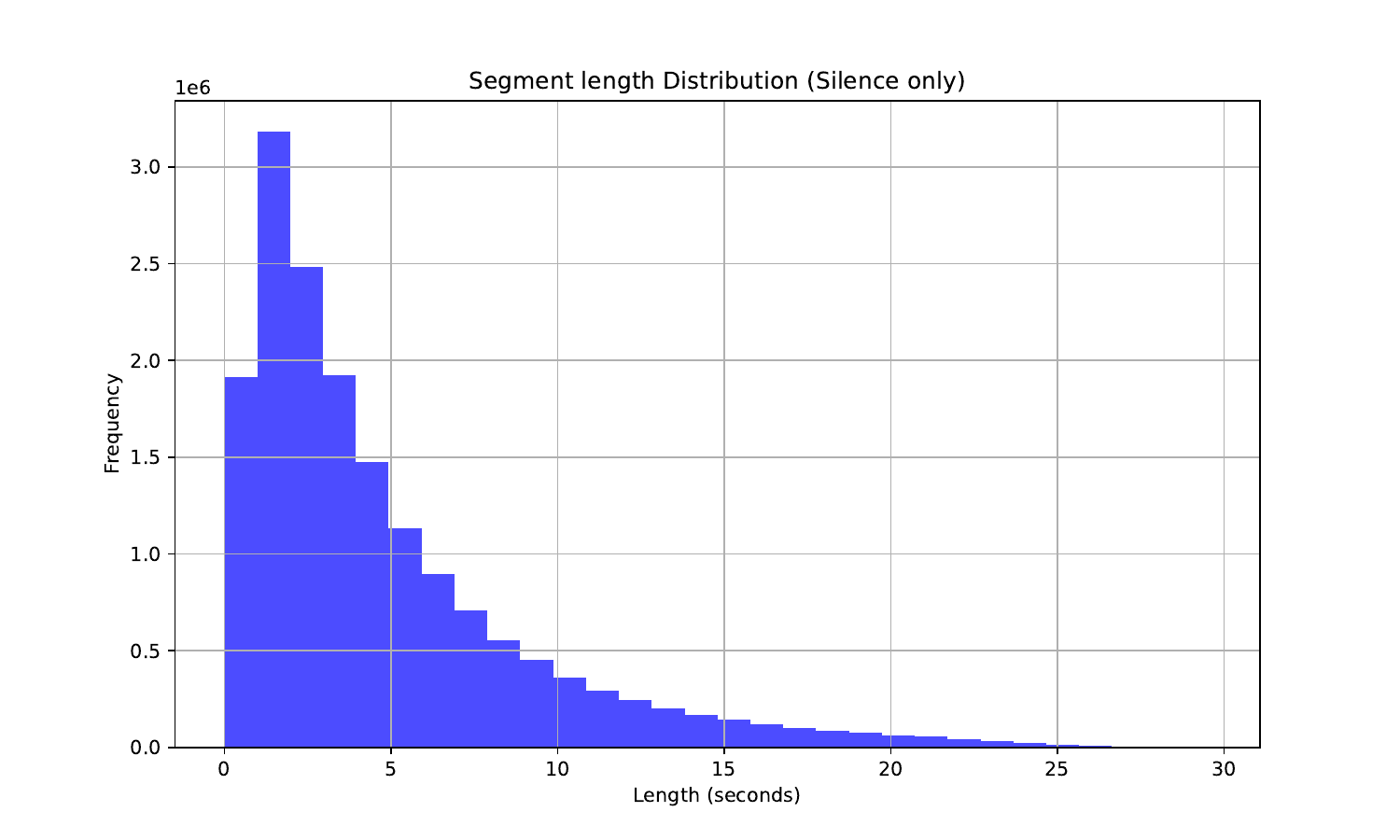}
        \caption{Silence}
        \label{fig:seg-il-silence}
    \end{subfigure}
    \hfill
    \begin{subfigure}{0.48\columnwidth}
        \centering
        \includegraphics[width=\linewidth]{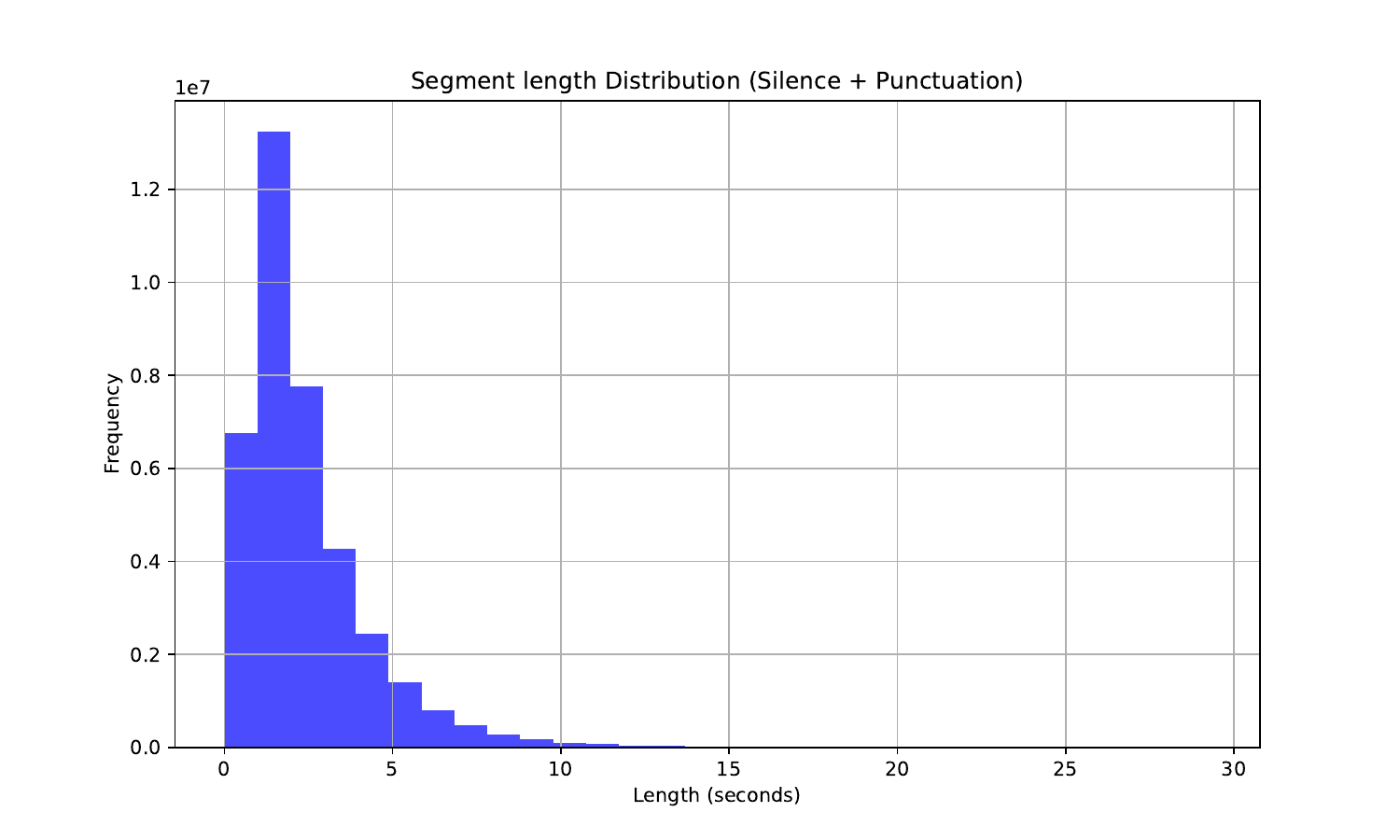}
        \caption{Silence + Punctuation}
        \label{fig:seg-il-silence-punc}
    \end{subfigure}
    \caption{Segment length distribution of the interleaved sequence using acoustic signal only (left) and using both acoustic and semantic clues (right).}
    \label{fig:two_images}
\end{figure}

Table~\ref{tab:ablation_segment} presents an ablation study on interleaved (IL) sequence construction. The table compares practical JSTIP recipes, so rows that add PubMed text, TTS-pairs, TTS-transcriptions, or interleaved variants should not be interpreted as isolating a single factor in all cases. The cleanest evidence for interleaving comes from comparisons within the same data condition, such as ASR-only versus +Interleave, ASR+PubMed versus +Interleave, and ASR+TTS-transcription versus +Interleave. Segment boundaries are derived from acoustic silence alone or from both silence and punctuation. All interleaving strategies outperform ASR-only training in entity accuracy and cross-modal generalization. Word-IL alone improves entity recognition but is less effective at reducing the modality gap, while Segment-IL greatly improves cross-modal consistency, showing the importance of longer-range semantic alignment between speech and text beyond individual words. The improvements are more significant when punctuation is incorporated for segmentation. We use punctuation because acoustic silence can be sparse in some utterances, leading to very long segments; punctuation creates a more balanced segment length distribution. Figure~\ref{fig:two_images} validates this assumption: segments are more concentrated in shorter regions when punctuation is used, making the supervision signal more balanced and facilitating modality alignment. Despite the greater modality-gap reduction from punctuation-aware segmentation, Segment-IL achieves similar EER to silence-only segmentation. Mixing Word-IL with Segment-IL transfers the modality-gap reduction into stronger entity improvements, indicating that word-level interleaving provides ASR-specific benefits beyond segment-level interleaving used in prior systems such as Voxtral~\cite{liu2025voxtral} and Xie et al.~\cite{xie2025enhancing}.

\section{Conclusion}
\label{sec:conclusion}

In this work, we revisit speech-LLM integration from the pretraining perspective and identify a key limitation of conventional approaches: decoder-only speech-LLM models tend to over-specialize in speech-conditioned decoding, leading to weak utilization of LLM knowledge and textual priors. To address this limitation, we propose Joint Speech-Text Interleaved Pre-Training (JSTIP), an ASR-oriented pretraining strategy that constructs interleaved speech-text sequences within aligned pairs to preserve the LLM's generative behavior during adaptation. The central novelty is to make text prediction occur after both acoustic and textual contexts, including scalable word-level interleaving for continuous speech representations and its combination with segment-level interleaving. Experiments on our internal ASR setup show that JSTIP consistently improves entity recognition, reduces the modality gap, and enables zero-shot SQA without additional task-specific supervision. We further show that domain transcription data becomes as effective as synthetic TTS pairs under interleaved training, revealing when and how LLM pretraining benefits ASR under large supervised ASR data. Because the main training data, domain evaluation sets, and backbone components are internal, the primary evidence should be interpreted as controlled improvements within our system rather than a universal ranking over Speech-LLMs.


\bibliographystyle{IEEEtran}
\bibliography{mybib}

\end{document}